\documentclass[runningheads]{llncs}
\usepackage[T1]{fontenc}
\usepackage{subcaption}
\usepackage{glossaries}
\usepackage{graphicx}
\begin{document}
%
\newacronym{ADAS}{ADAS}{Advanced Driving Assistant Systems}
\newacronym{LKAS}{LKAS}{Lane Keeping Assistance System}
\newacronym{LKA}{LKA}{Lane Keeping Assist}
\newacronym{LDW}{LDW}{Lane departure warning}
\newacronym{LCA}{LCA}{Lane Change Assistant}

\newacronym{ALC}{ALC}{Adaptive Light Control}
\newacronym{ACC}{ACC}{Adaptive Cruise Control}
\newacronym{FCW}{FCW}{Forward Collision Warning}

\newacronym{ADS}{ADS}{Automated Driving Systems}
\newacronym{CNN}{CNN}{Convolutional Neural Network }
\newacronym{ROS}{ROS}{Robot Operating System}
\newacronym{PCL}{PCL}{Point Cloud Library}
\newacronym{SAE}{SAE}{Society of Automotive Engineers}
\newacronym{DDT}{DDT}{Dynamic Driving Task}
\newacronym{ODD}{ODD}{Operational Design Domain}
\newacronym{LDS}{LDS}{Laser Distance Sensor}
\newacronym{LIDAR}{LIDAR}{Light Detection And Ranging}

\newacronym{IMU}{IMU}{Inertial Measurement Unit }
\newacronym{GPS}{GPS}{Global Positioning System}
\newacronym{GNSS}{GNSS}{Global Navigation Satellite System}
\newacronym{PPS}{PPS}{Pulse per Second}
\newacronym{FoV}{FoV}{Field of View}
\newacronym{Gbps}{Gbps}{Gigabits per second}
\newacronym{OBDII}{OBDII}{On-board Diagnostics II}
\newacronym{CAN}{CAN}{ Controller Area Network}
\newacronym{PoE}{PoE}{Power over Ethernet}
\newacronym{AR}{AR}{Augmented Reality}
\newacronym{SSD}{SSD}{Solid State Drive}

\title{JKU-ITS Automobile for Research on Autonomous Vehicles\thanks{This work was partially supported by the Austrian Ministry for Climate Action, Environment, Energy, Mobility, Innovation and Technology (BMK) Endowed Professorship for Sustainable Transport Logistics 4.0., IAV France S.A.S.U., IAV GmbH, Austrian Post AG and the UAS Technikum Wien. It was additionally supported by the Austrian Science Fund (FWF), project number P 34485-N.}}
%
%

\author{Novel Certad\inst{1}\orcidID{0000-0002-0681-7658} \and
Walter Morales-Alvarez\inst{1}\orcidID{0000-0001-6912-4130} \and
Georg Novotny\inst{1,2}\orcidID{0000-0001-8990-2622} \and
Cristina Olaverri-Monreal\inst{1}\orcidID{0000-0002-5211-3598]}}

\authorrunning{N. Certad et al.}
%

\institute{Chair for Sustainable Transport Logistics 4.0,\\ Johannes Kepler University Linz, Austria\\
\email{\{novel.certad\_hernandez,walter.morales\_alvarez,\\cristina.olaverri-monreal\}@jku.at}\\
\url{www.jku.at/its} \and
UAS Technikum Wien, H\"ochstaedtplatz 6, 1200  Vienna, Austria
\email{georg.novotny@technikum-wien.at}}

\maketitle              
\begin{abstract}
In this paper, we present our brand-new platform for Automated Driving research. The chosen vehicle is a RAV4 hybrid SUV from TOYOTA provided with exteroceptive sensors such as a multilayer LIDAR, a monocular camera, Radar and GPS; and proprioceptive sensors such as encoders and a 9-DOF IMU. These sensors are integrated in the vehicle via a main computer running ROS1 under Linux 20.04. Additionally, we installed an open-source ADAS called Comma Two, that runs Openpilot to control the vehicle. The platform is currently being used to research in the field of autonomous vehicles, human and autonomous vehicles interaction, human factors and energy consumption. 
\end{abstract}
\section{Introduction}
Developing an automated vehicle for research is a difficult task that requires effort, knowledge, and most of the time a big budget. The first milestone to achieve this automation is to provide the vehicle with drive-by-wire capabilities that allow the developers to control its actuators with digital signals. Most of the research platforms that can be found in the literature relied on complex adaptations of expensive and sometimes unreliable electro-mechanical structures to steer the vehicle and actuate over the throttle and brake pedals.
Nowadays, a lot of commercial vehicles, with  \gls{SAE}, level 2 functionalities, interact with the actuators using on-board computers that communicate through a CAN bus. Although it is technically possible to access the CAN through an available port like the \gls{OBDII}, the encoding and decoding process of the CAN frames relies on the availability of a proprietary database owned by the automotive company that developed the car. As a result, there is a growing community of automotive enthusiasts, constantly reverse-engineering CAN databases, allowing users to gain partial control of some vehicles.
In this paper, different possibilities to achieve drive-by-wire capabilities in a TOYOTA RAV4 hybrid SUV were considered. The final decision was taken under the following premises:  

\begin{itemize}
    \item The vehicle has to maintain all the original functionalities and driving capabilities. 
    \item Low or no intervention at all in the original structure of the vehicle. 
    \item It has to be possible for a human driver to take over the control of the vehicle instantly at any moment.
    \item Fast development and integration (less than six months).
    \item Low cost (less than twenty thousand Euros).
\end{itemize}

While lots of papers focus on perception, control, planning, mapping, decision-making, and other autonomous vehicle software-related technologies, the hardware and physical architecture design are rarely discussed \cite{demiguel2020}. In this paper, we will present the placement of the sensors, power management system, and data management implemented in the vehicle along with the software architecture of the vehicle.

The  remainder  of  this paper  is  organized  as  follows: the section~\ref{sec:hardware} describes the sensors and physical architecture design;  section~\ref{sec:software} details all the software implemented; section~\ref{sec:results}  presents  some experiments already carried out using the vehicle. Finally section~\ref{sec:conclusion} concludes the present study outlining future research.

\section{Hardware Setup}\label{sec:hardware}
\begin{figure}[t]
    \centering
    \includegraphics[width=0.7\textwidth]{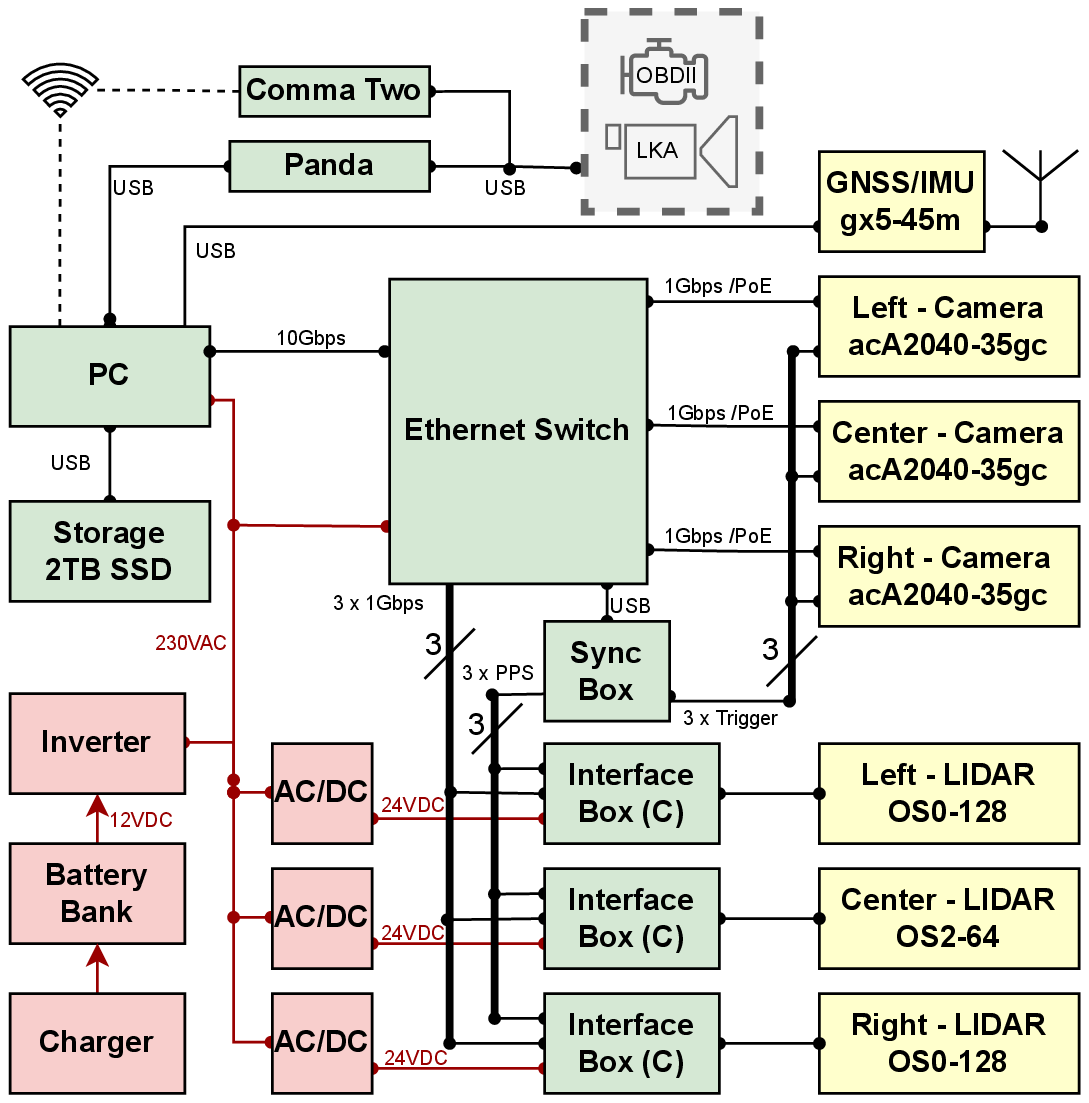}
    \caption{This diagram depicts the connection between all the modules and sensors available in the vehicle. Sensors are shown in yellow, green is used to depict the processing \& communication components, and the power management module can be seen in red.} \label{fig:hardware}
\end{figure}
As stated above, the vehicle used is a RAV4 hybrid SUV from TOYOTA. The additional hardware is divided into three modules according to their main functionalities:

\begin{itemize}
    \item \textbf{Processing and communication:} it contains the processing unit and storage along with all the necessary network infrastructure for communication between sensors and the processing unit, synchronization of the acquisition, and reading the car's data. The components of the module are colored green in Fig.~\ref{fig:hardware}.
    \item \textbf{Sensors:} it consists of the \gls{LIDAR} sensors, cameras, \gls{IMU} and \gls{GNSS} as can be seen in Fig.~\ref{fig:hardware} colored yellow.
    \item \textbf{Power management:} This module is in charge of providing power to the other two modules. These components are colored red in Fig.~\ref{fig:hardware}.
\end{itemize}

\subsection{Sensors}

The JKU-ITS Automobile is equipped with different sensors to perceive the surrounding environment and locate the vehicle within it.

\subsubsection{LIDAR Sensors:}

\begin{figure}[ht]
	\centering
	\begin{subfigure}{0.6\textwidth}
		\includegraphics[width=\textwidth]{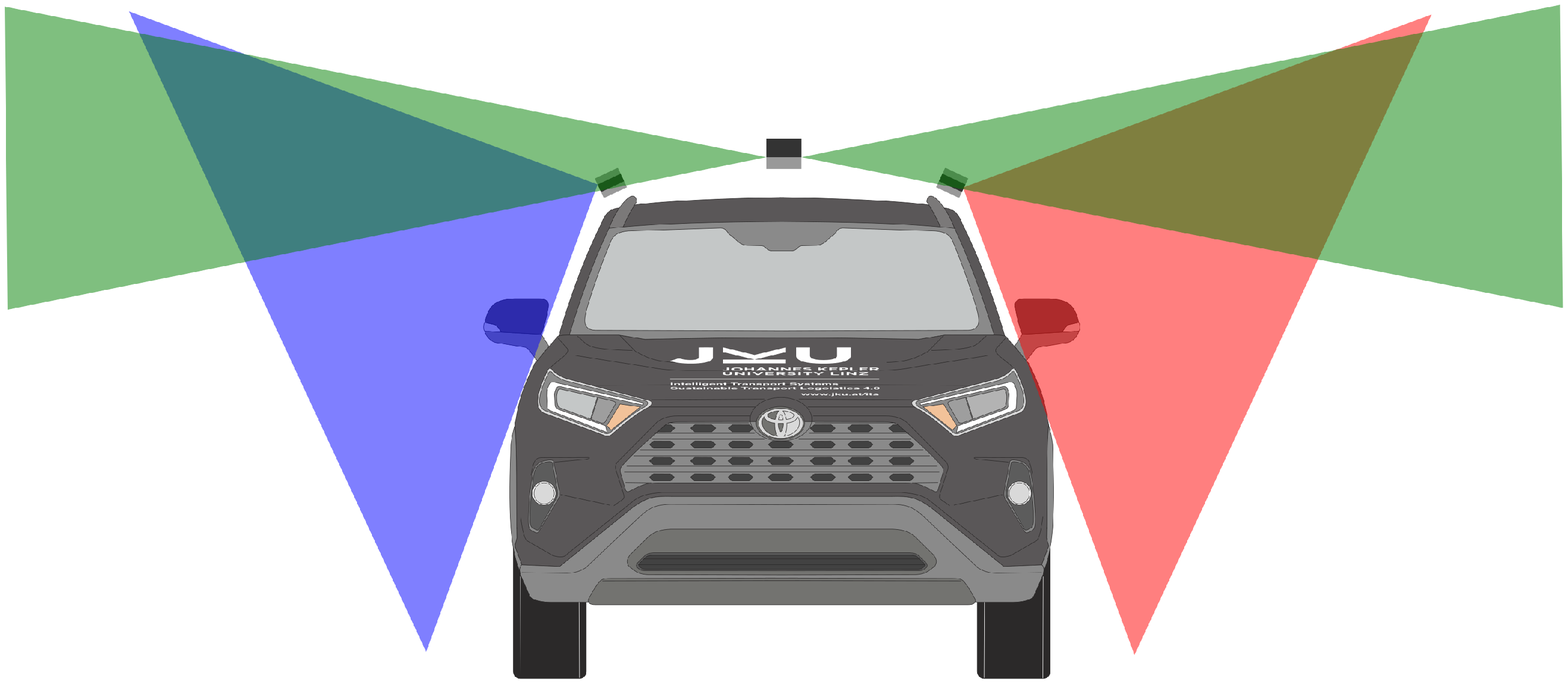}
		\caption{}
	    \label{fig:lvfov}
	\end{subfigure}
	\\
	\begin{subfigure}{0.4\textwidth}
		\includegraphics[width=\textwidth]{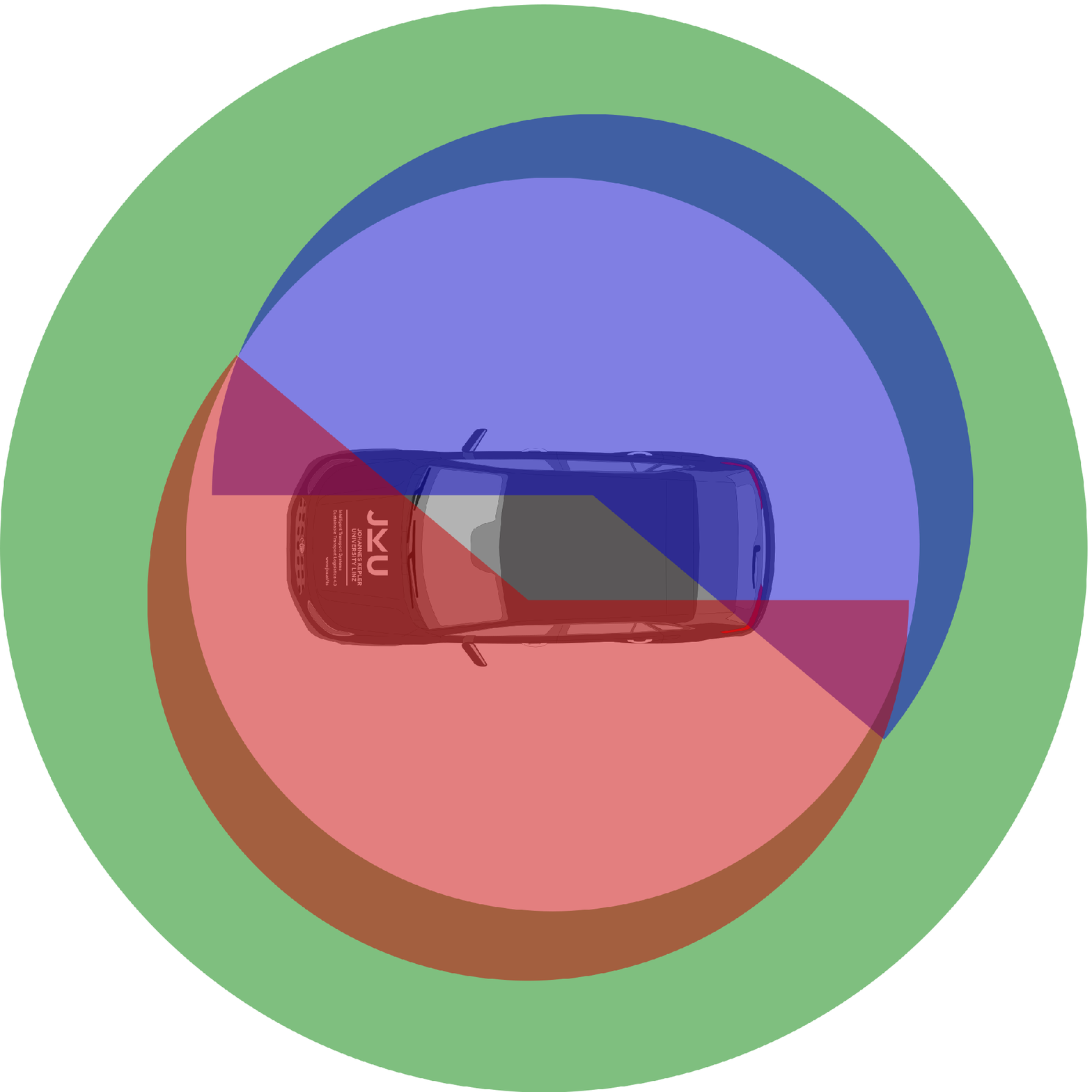}
		\caption{}
	    \label{fig:lhfov}
	\end{subfigure}
    \begin{subfigure}{0.4\textwidth}
		\includegraphics[width=0.65\textwidth]{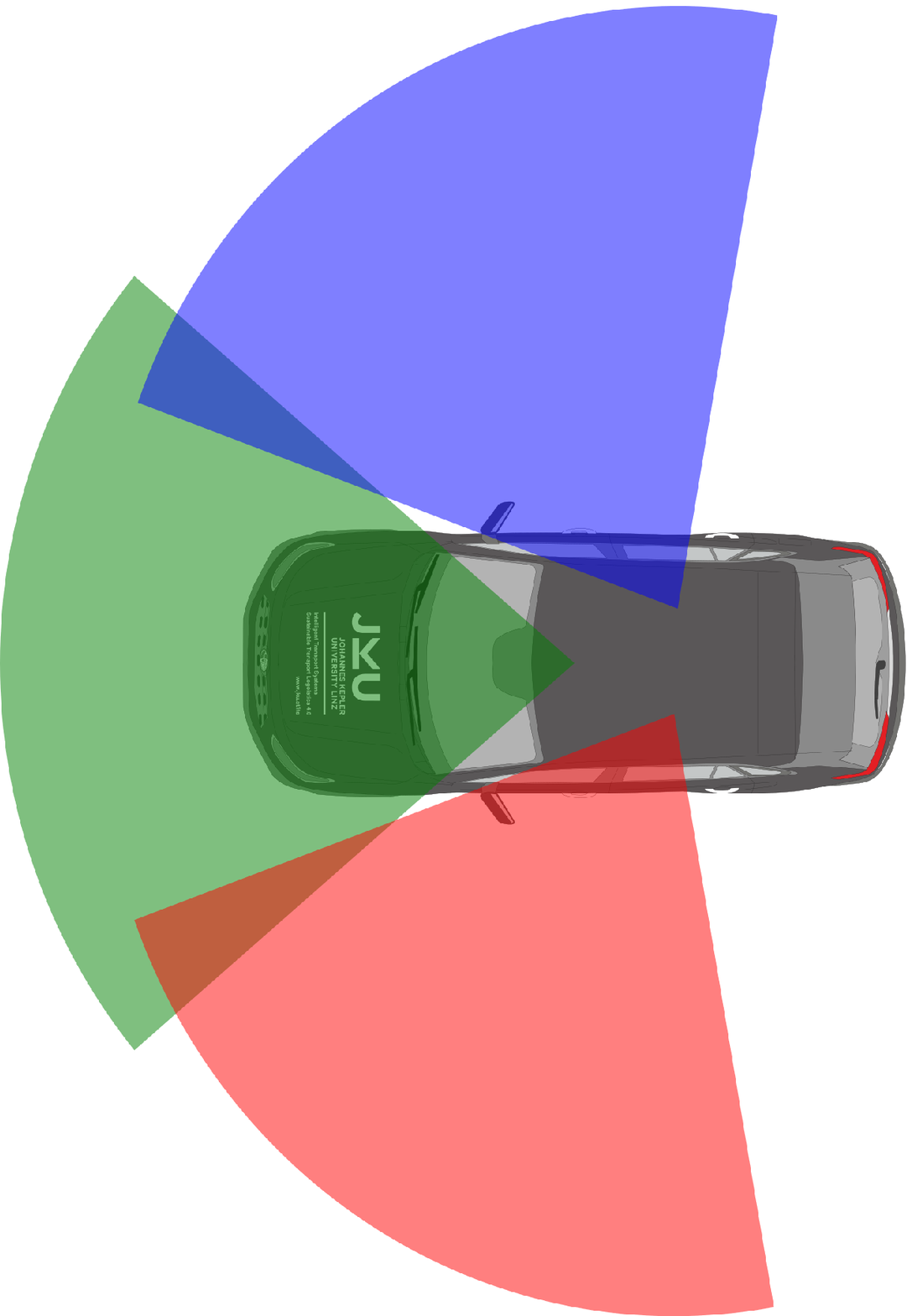}
		\caption{}
	    \label{fig:chfov}
	\end{subfigure}
	\caption{(a) depicts the location of the three \gls{LIDAR} sensors and the vertical \gls{FoV} associated with each one of them. (b) shows the horizontal \gls{FoV} of the \gls{LIDAR} sensors. Lastly, (c) depicts the horizontal \gls{FoV} covered by the three cameras.}
	\label{fig:fov}
\end{figure}

The vehicle is equipped with three different \gls{LIDAR}s. An OS2 (range from 1 to 240m, $22.5^\circ$ of vertical \gls{FoV}, and 64 layers) is in the center of the roof. This sensor has a long range of view, covering objects far away from the car (Fig.~\ref{fig:lvfov}). Two OS0 (range from 0.3m to 50m, $90.0^\circ$ of Vertical \gls{FoV}, and 128 layers) cover the objects in the direct vicinity of the vehicle. One is in the front-left part of the roof, and the other is in the rear-right part. As seen in Fig.~\ref{fig:lvfov}, both OS0 sensors are inclined $22.5^\circ$ towards the plane to cover the sides of the vehicle. This configuration with three \gls{LIDAR} sensors allows us to cover the whole area surrounding the car without noticeable dead zones (Fig.~\ref{fig:lhfov}).

\subsubsection{Cameras:}
Three acA2040-35gc Basler GigE cameras (3.2MP, Angle of View: $79.0^\circ$ horizontal, $59.4^\circ$ vertical) are located on the roof, covering a horizontal field of view of approximately $200^\circ$, as seen in Fig.~\ref{fig:chfov}. The cameras are configured to acquire 10 frames per second according to an external 10Hz signal (trigger) generated by the synchronization box.

\subsubsection{GNSS INS:}
The \gls{GNSS} receiver (3DMGX5-45) is located underneath the center-\gls{LIDAR}. It has a maximum refreshing rate of 4Hz, and provides the raw \gls{GPS} data, raw \gls{IMU} data, and Kalman filtered odometry data. The \gls{GNSS} receiver is also connected to an external antenna. The device is connected directly to the processing unit via USB.

\subsection{Processing \& Communication}
This module is in charge of connecting every sensor to the processing unit.

\subsubsection{Processing Unit:}
The processing unit varies according to the configuration of the vehicle. Most of the time, it is only utilized for data acquisition, and the processing is done afterwards. The base setup uses a Mini PC (Intel Core i3, 8GB RAM) and a high-speed (up to 1050MB/s) external \gls{SSD} for data storage.

\subsubsection{Network switch:}
The core of the communication module is the network switch (GS728TXP - Netgear, 28 x 1Gbit, 4 x 10Gbit ports, \gls{PoE}). The three cameras and the three \gls{LIDAR}s are connected to the same network using six 1\gls{Gbps} ports. The Processing unit is connected using a 10\gls{Gbps} port. The required bandwidth was calculated to ensure the system works without losing packages. The required bandwidth to acquire images at 10Hz was 30MB/s per camera ($BW_{cameras}$). Similarly, to acquire pointclouds at 10Hz using the OS2 \gls{LIDAR}, 30MB/s bandwidth is required ($BW_{OS0}$). The OS0 \gls{LIDAR}s have twice the number of layers compared to the OS2, thus the required bandwidth is 60MB/s ($BW_{OS2}$). Considering only the \gls{LIDAR}s and cameras, the total bandwidth ($BW_{Total}$) was calculated by applying the following formulas:

 \begin{equation}\label{equ:bandwidth}
      BW_{Total} = 3 \times BW_{cameras} + 2 \times BW_{OS0} + BW_{OS2}\\
 \end{equation}
 \begin{equation}
      BW_{Total} = 3 \times 30MB/s + 2 \times 60MB/s  + 30MB/s = 240MB/s
 \end{equation}
 \begin{equation}
      BW_{Total} = 1.92Gbps
 \end{equation}

The data streams produced by the \gls{GNSS}/\gls{IMU}, the car's state read from the Black Panda, and the overhead created by the \gls{ROS} headers are negligible compared to the \gls{LIDAR}s and cameras. The real total bandwidth ($BW_{Total}$) is expected to be a little bigger considering these three small data streams. However, using a 10\gls{Gbps} link between the network switch and the processing unit provides about 4 times the required bandwidth.

\subsubsection{Sync Box:}
The Synchronization Box is a microcontroller unit that generates three 10Hz signals to trigger each one of the cameras, and three PPS signals at 1Hz to synchronize the \gls{LIDAR} frames acquisition.

\subsubsection{Black Panda \& Comma Two:}
Both devices were developed by \textit{comma.ai}\cite{comma}. Black Panda is a universal car interface that provides access to the communication buses of the car through the \gls{OBDII} port. Comma Two is an embedded system designed to run \textit{Openpilot} (see section \ref{sec:ads}). It has a front camera to perceive the road and a back camera for driver monitoring, including infrared LEDs for night. A Black Panda is integrated within the Comma Two, allowing the connection to the \gls{OBDII} port and the stock \gls{LKA}'s camera of the vehicle.

Only one of the devices can be connected (to the \gls{OBDII} port) at the same time. When acquiring data without the need for Advanced Driving Assistant Systems \gls{ADAS}, we use the Panda device. Otherwise, we operate the \textit{Openpilot} software with Comma Two. The connection between the Black Panda and the processing unit is established via USB-C, while Comma Two uses a WIFI network to connect to the processing unit. 

Drive-by-wire is enabled by sending commands to the vehicle through either device.

\subsection{Power management}

Since the vehicle is hybrid, it was originally equipped with an auxiliary battery used to power the vehicle's electrical system (as in a conventional vehicle) and a battery pack (244.8V) for powering the electrical motors. To avoid damage to the original vehicle equipment, we designed an independent power system for the sensors. The system is powered by five 12VDC-120Ah batteries connected to a pure sine-wave inverter able to produce 230VAC. A power outlet is connected to the inverter to distribute the voltage between the \gls{LIDAR}s, the network switch, and the computers. The cameras are powered via \gls{PoE} fed by the network switch. There is also a battery charger connected to the batteries. It relies on a conventional 230VAC input.

\section{Software Architecture}\label{sec:software}

The software is almost completely based on \gls{ROS}\cite{ROS}. We developed a custom package to launch the ROS-driver for each sensor and the associated transformations. The drivers used by the sensors were:

\begin{itemize}
    \item \textit{microstrain\_mips}: is the \gls{ROS} driver for microstrain \gls{IMU}. The \textit{microstrain.launch} file is used to start the \gls{GNSS}/\gls{IMU} stream.
    \item \textit{ pylon-ROS-camera}: is the \gls{ROS} driver supplied by Basler to be used with the Ethernet cameras.
    \item \textit{ouster-lidar/ouster\_example}: is the \gls{ROS} driver to start the sensor-stream of the three \gls{LIDAR}s.
\end{itemize}

\subsubsection{Sensor Calibration}
The intrinsic calibration of the cameras was achieved using the standard \gls{ROS} package \textit{camera\_calibration}. The extrinsic calibration was made following the automatic procedure described in~\cite{beltran2022} with the available \gls{ROS} package. The three cameras and the two side-\gls{LIDAR}s were calibrated using the center-\gls{LIDAR} location as the reference frame. 

\subsubsection{\gls{ADS}}\label{sec:ads}

The JKU-ITS automobile achieved \gls{SAE} level 3 under certain conditions, that enabled the vehicle to drive in automated mode until a take over request was triggered. To this end, the vehicle relies on an open-source \gls{ADS} developed to run on the Comma Two device, called Openpilot \cite{openpilot}. Openpilot runs several \gls{ADAS} within the same embedded hardware including: \gls{LKA}, \gls{ALC}, \gls{FCW}, \gls{ACC}, and \gls{LDW}. 

We also developed a custom bridge that allowed us to connect \gls{ROS} with the Openpilot's message manager (cereal). The bridge was written in python. When the Black Panda is connected, the car's state can be read from the \gls{OBDII} port. Drive-by-wire is also possible using this bridge. When the Comma Two is connected, the bridge also allows us to send and receive messages to or from Openpilot (e.g. engage or disengage the \gls{ADAS}).

\section{Results}\label{sec:results}

\begin{figure}[ht]
    \centering
    \includegraphics[width=0.6\textwidth]{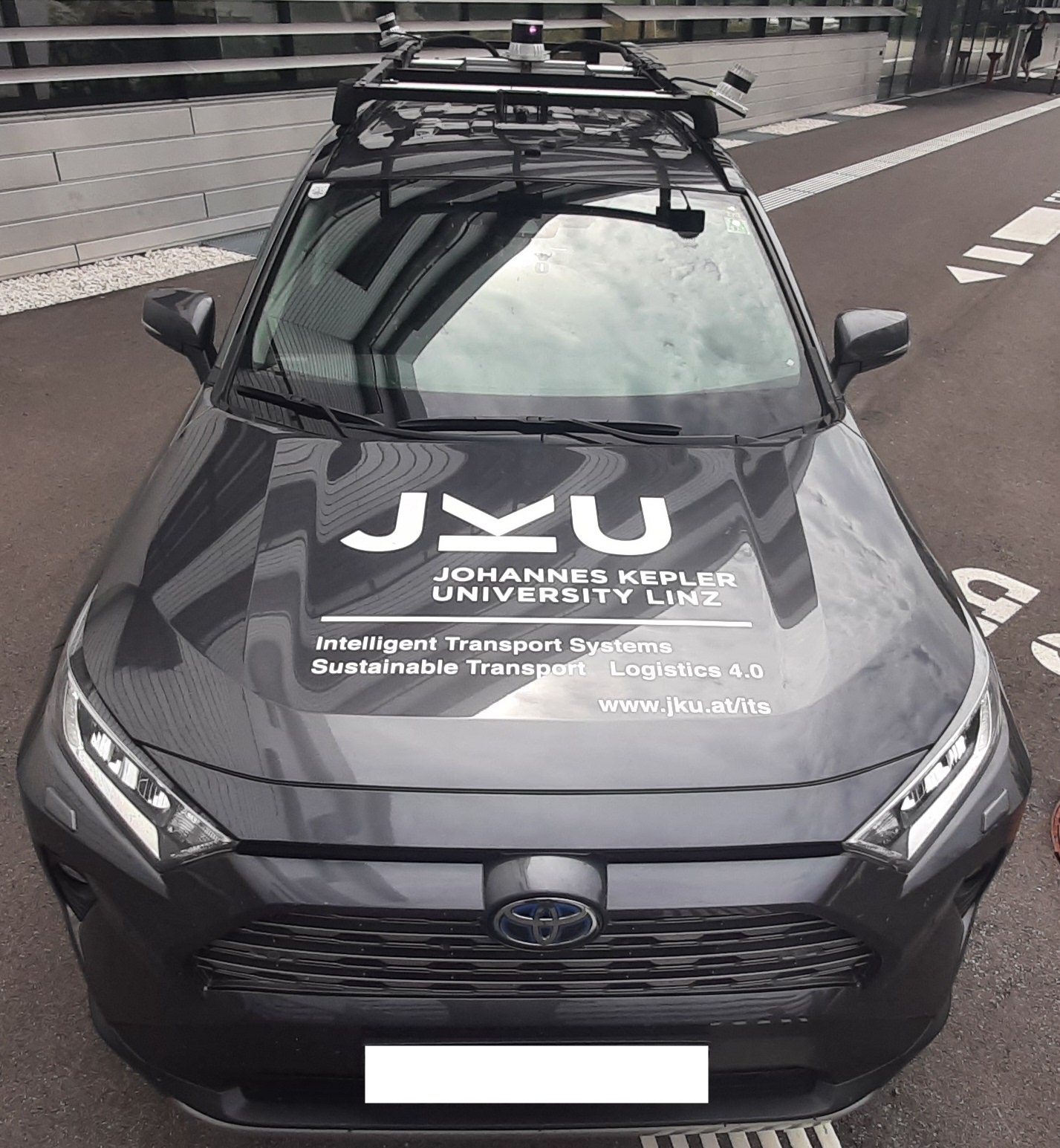}
    \caption{JKU-ITS Automobile picture showing the LIDAR sensors and cameras mounted over the roof rack} \label{fig:vehicle}
\end{figure}

The resulting platform with the sensors is shown in Fig.~\ref{fig:vehicle}. It has been used in several demonstrations and also has been central to two already published projects. In \cite{validi2021}, the battery energy estimation model in SUMO (Simulation of Urban Mobility) was compared with the battery energy consumption of the vehicle. It was found that there exist big differences between the estimated model and the real data. Then in \cite{morales2021}, the response of drivers to a take-over request at different driving speeds while being engaged in non-driving-related tasks, was studied. In addition, further experiments are been conducted with a haptic guidance system to avoid sudden obstacles on the road \cite{morales2022}.

\section{Conclusion \& Future work}\label{sec:conclusion}

The JKU-ITS Automobile is a modular platform that was developed to perform research on autonomous vehicles at the ITS-Chair Sustainable Transport Logistics 4.0. at the Johannes Kepler University Linz in Austria.

The vehicle is fully functional and geared with the required sensors, and further implementations are being conducted. Three more cameras will soon cover $360^\circ$ \gls{FoV} around the vehicle. We also plan to assemble a custom connector to extract the \gls{PPS} signal from the \gls{GNSS} and feed it to the Synchronization Box. The migration of the software from \gls{ROS}1 to \gls{ROS}2 is already been implemented.


%
%
%
\bibliographystyle{splncs04}
\bibliography{mybib}
\end{document}